\documentclass[letterpaper]{article} 
\usepackage{aaai24}  
\usepackage{times}  
\usepackage{helvet}  
\usepackage{courier}  
\usepackage[hyphens]{url}  
\usepackage{graphicx} 
\usepackage{natbib}  
\usepackage{caption} 
\usepackage{algorithm}

\usepackage{amsmath}
\usepackage{amsfonts,amssymb}
\usepackage{multirow}
\usepackage{booktabs}
\usepackage{bbding}
\usepackage{epsfig}
\usepackage{diagbox}
\usepackage{algpseudocode}
\usepackage{newfloat}
\usepackage{listings}
\DeclareCaptionStyle{ruled}{labelfont=normalfont,labelsep=colon,strut=off} 
\floatstyle{ruled}
\newfloat{listing}{tb}{lst}{}
\floatname{listing}{Listing}
\title{TCNet: Continuous Sign Language Recognition from\\ Trajectories and Correlated Regions}
\author{
    Hui Lu, Albert Ali Salah, Ronald Poppe
}
\affiliations{
    Utrecht University\\
    h.lu1@uu.nl, a.a.salah@uu.nl, r.w.poppe@uu.nl\\

}

\usepackage{bibentry}

\begin{document}

\maketitle

\begin{abstract}
A key challenge in continuous sign language recognition (CSLR) is to efficiently capture long-range spatial interactions over time from the video input. To address this challenge, we propose TCNet, a hybrid network that effectively models spatio-temporal information from Trajectories and Correlated regions. TCNet's trajectory module transforms frames into aligned trajectories composed of continuous visual tokens. In addition, for a query token, self-attention is learned along the trajectory. As such, our network can also focus on fine-grained spatio-temporal patterns, such as finger movements, of a specific region in motion. TCNet's correlation module uses a novel dynamic attention mechanism that filters out irrelevant frame regions. Additionally, it assigns dynamic key-value tokens from correlated regions to each query. Both innovations significantly reduce the computation cost and memory. We perform experiments on four large-scale datasets: PHOENIX14, PHOENIX14-T, CSL, and CSL-Daily, respectively. Our results demonstrate that TCNet consistently achieves state-of-the-art performance. For example, we improve over the previous state-of-the-art by 1.5\% and 1.0\% word error rate on PHOENIX14 and PHOENIX14-T, respectively. Code is available at https://github.com/hotfinda/TCNet.
\end{abstract}

\section{Introduction}
Continuous sign language recognition (CSLR) is the task of transcribing sequences of continuous gestures made by a signer into sentences. Unlike isolated sign language recognition, CSLR algorithms are typically trained in a weakly-supervised way because only sentence-level labels are provided. Temporal information is only available implicitly. One challenge is to extract relevant patterns over time. 

Traditionally, CSLR has been addressed by first extracting spatial patterns from video, and subsequently considering temporal relations between these features, typically using LSTMs (e.g., \citet{min2021visual,niu2020stochastic}). Since sign language is mainly understood from signers’ faces and hands, some CSLR models~\cite{papadimitriou2020multimodal,zhou2020spatial} leverage pre-trained pose detectors to locate the face and hands and then crop the feature maps to form a multi-stream architecture. This approach allows to focus on the potentially relevant regions in the video input, but is unsuited to exploit interactions between these regions.

\begin{figure}[t]
    \begin{center}
    \includegraphics[width=\linewidth]{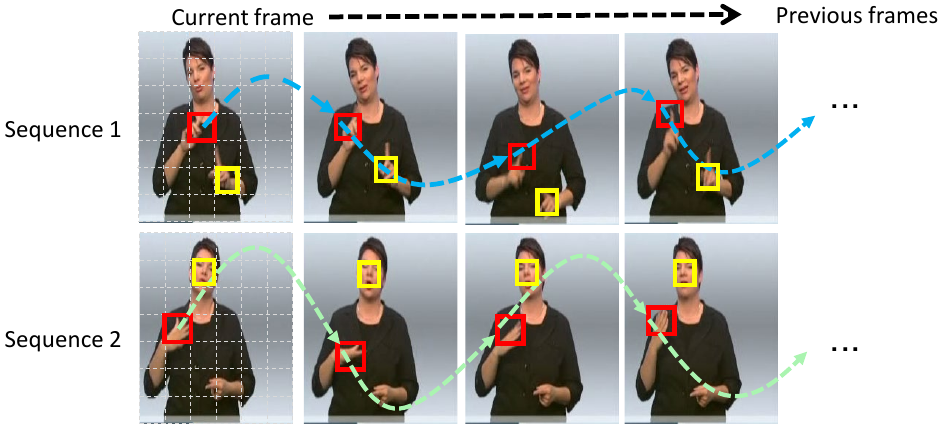}
    \caption{Trajectories and correlated regions. The sign is revealed by the trajectories of both hands in Seq. 1, and the right hand and head in Seq. 2. The content of these regions, such as the hand pose, is also important. We visualize the trajectory of the right hand (red) and indicate correlated regions (yellow).}
    \label{fig:1}
    \end{center}
\end{figure}

More recently, transformer models allow focusing on informative spatial regions in the input by employing self-attention~\cite{hu2023continuous,hu2023self}. Although self-attention has been shown to improve recognition performance, its high computational cost limits its use to narrow spatial windows. Moreover, required pairwise token affinity computations across all spatial locations amount to a high computational complexity and incur significant memory footprints. 

For the CSLR task, correlated regions to express sign language are generally dynamic. In Figure~\ref{fig:1}, the hand regions are correlated in Seq.1, but not in Seq.2. This requires us to assign dynamic key-value tokens from correlated regions to each query, such that irrelevant tokens can be avoided.
Moreover, extracting temporal information from correlated regions is also important to understand the sign language.

We build on these notions by explicitly focusing on extracting trajectories, considering patterns along these trajectories, and modeling the correlations between relevant regions, without making any assumptions about these regions. We propose a novel network named TCNet, which collects temporal information over an entire sequence and attends to correlated regions to collect spatial information.

Our main innovations are the trajectory and correlation modules. The trajectory module transforms temporal movements into pre-aligned trajectories of visual tokens, without relying on face or hand detection. Additionally, self-attention is calculated along these trajectories. The correlation module calculates attention per frame, where only the correlated regions are considered. 
TCNet can effectively extract spatial-temporal information with significantly reduced computation cost and memory compared to related work, while showing improved performance.

In this paper, we make the following contributions: 
\begin{itemize}
\item We propose TCNet, a hybrid network for CSLR that employs innovative trajectory and correlation modules in the feature extractor to improve the feature representation.

\item Both modules are combined to yield an effective network with reduced computation and memory cost.

\item TCNet achieves state-of-the-art results on PHOENIX14, PHOENIX14-T, CSL, and CSL-Daily. Our ablation studies demonstrate the added value of trajectory and correlation modules. Experiments on various backbones demonstrate consistent performance gains.

\item We use standard experimental protocols and release all our code for reproducibility.
\end{itemize}

\section{Related Work}
\label{sec:related work}

The goal of CSLR is to translate sequences of image frames into corresponding glosses, the atomic lexical units of sign language. The task is weakly supervised because only sentence-level labels are provided. Hence, there is a challenge in extracting relevant spatio-temporal patterns from the input frames. Recent progress in CSLR is mainly due to advances in deep learning. We discuss current work in extracting spatial and temporal patterns, respectively.

\subsection{Extracting Spatial Patterns}
Several methods~\cite{min2021visual,niu2020stochastic} employ 2D CNN models to extract frame-wise features, and adopt LSTMs for long-term temporal modeling. 
Since the most informative regions in the video input correspond to the face and hands~\cite{koller2020quantitative}, some CSLR models~\cite{papadimitriou2020multimodal,zhou2020spatial, zhou2021spatial} leverage pre-trained face/hand detectors or pose estimation algorithms (e.g., \citet{cao2021openpose,sun2019deep}) to localize relevant parts of the body. The corresponding regions are cropped from the input and used as additional inputs in multi-stream architectures. This strategy allows to focus on relevant parts of the input but interactions between the regions cannot be readily used~\cite{zuo2022c2slr}.

Inspired by recent progress in transformers, researchers have used self-attention for CSLR to enhance the feature extractor, by focusing on relations between regions within a single frame \cite{hu2023continuous,hu2023self}. Although self-attention has been shown to improve recognition performance since the receptive field is increased, the high computational cost restricts it use to limited spatial windows. Moreover, there is a significant computation cost involved in calculating the pairwise token affinity between all potential spatial locations.

To address this issue, researchers have proposed attention variants to extract correlated regions and to reduce the computation cost. For example, \citet{zuo2022c2slr} use a lightweight spatial attention module guided by keypoint heatmaps to enforce a focus on informative regions. 

We argue that, specifically for the CSLR task, not all regions in the frame contribute equally to recognition since the sign language are mainly conveyed through head and hand regions, Moreover, the correlated regions are generally dynamic. There is a need for mechanisms to attend to regions that can dynamically vary over time.

\subsection{Exploiting Temporal Information}
Recently, there is an increased focus on extracting temporal information \cite{hu2023continuous,hu2023self}. By leveraging the cross-frame temporal movements to express a sign, temporal patterns, for example of the hand and face, can be learned \cite{hu2023continuous}. Current approaches mainly use 3D convolutions~\cite{pu2019iterative} or their (2+1)D variants \cite{tran2018closer} to capture short-term temporal information.

More flexible ways of processing the temporal information, such as Temporal shift~\cite{lin2019tsm} or temporal convolutions~\cite{liu2020teinet}, also focus on short-term temporal movements. Recently, researchers have used adjacent frames to extract temporal information. ~\citet{hu2023self} extract temporal features from the difference between adjacent frames as approximate motion information, and then concatenate it with appearance features as input to self-attention, such that the network can capture spatio-temporal information. Similarly, CorrNet~\cite{hu2023continuous} uses a self-attention mechanism in temporally adjacent frames to collect body movement features. However, these approaches are limited to adjacent frames thus can only capture short-term temporal information.

Relevant movements can be temporally more or less distant. However, aggregating distinctive information from distant regions remains a challenge due to the limited spatial-temporal receptive field. To model variations, \citet{guo2023distilling} propose a dual-pathway network where two branches accommodate the local and global temporal context, respectively. We deviate from this approach by elegantly attending to dynamically varying regions, as are typically in sign language due to the continuous movement of the hands.

\begin{figure}[htb]
    \begin{center}
    \includegraphics[width=\linewidth]{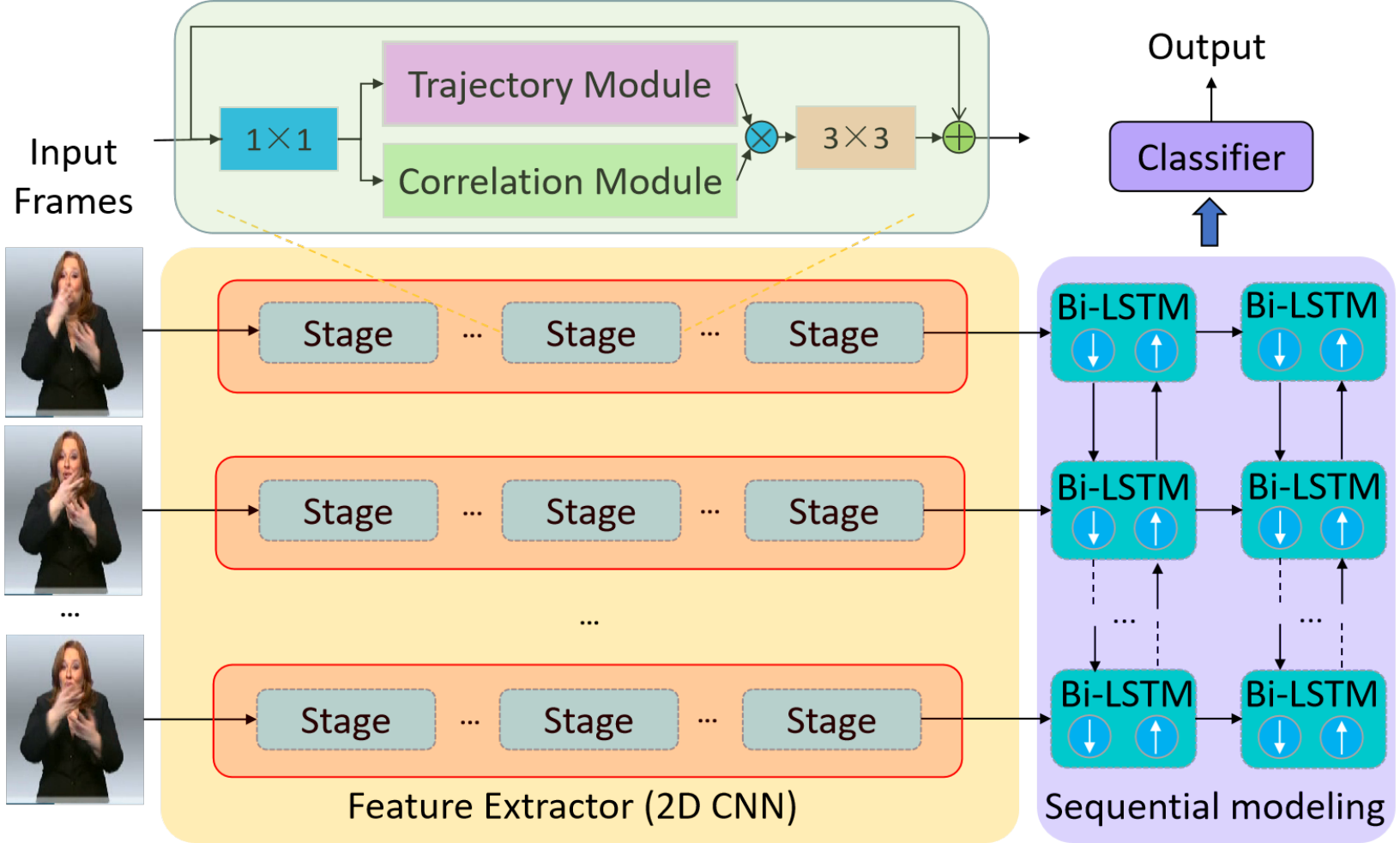}
    \caption{TCNet architecture with feature extractor, sequential modeling and classifier. TCNet blocks with our trajectory and correlation modules extract spatio-temporal features at various sequential stages.}
    \label{fig:2}
    \end{center}
\end{figure}

\section{Method}
\label{sec:method}
We introduce TCNet, a hybrid CNN-attention network that is capable of focusing on dynamically changing informative regions. To this end, we propose TCNet blocks that combine novel trajectory and correlation modules to enhance the spatio-temporal feature extraction capabilities.

\subsection{Overall Architecture}
The architecture of TCNet is shown in Figure~\ref{fig:2}. TCNet is based on previous works, including TLP~\cite{hu2022temporal} and CorrNet~\cite{hu2023continuous}, as we use the same feature extraction backbone, sequential modeling and classifier. This allows us to make pairwise comparisons to demonstrate the merits of the novel modules in the feature extraction module. But the TCNet block can be incorporated in a range of backbones, as we demonstrate in the ablation studies.

Continuous sign language recognition deals with a sequence mapping from a sign language video with $T$ frames $x = \left\{x_t \in \mathbb{R}^{w \times h \times c} \right\} = \left\{x_t\right\}_{t=1}^T$ to a sequence $y = \left\{y_i\right\}_{i=1}^L$ of length $L$, where $w \times h$ is the size of an input frame $x_t$, and the number of channels $c$ is three for RGB inputs. Our CSLR model consists of a feature extractor, a sequential modeling module and a classifier (see Figure~\ref{fig:2}). The feature extractor first processes input frames into fixed-length frame-wise features, $v = \left\{v_t \right\}_{t=1}^T \in \mathbb{R}^{T \times d}$, with $d$ the number of neurons in the last fully connected layer of the backbone. In case of a ResNet-18 backbone, $d = 512$. The processing is sequential and follows several stages, in line with previous work \cite{hu2022temporal,hu2023continuous}. Details of the network architecture appear in the supplementary material. 
After feature extraction, a series of 2-layer Bi-directional Long Short-Term Memory (Bi-LSTMs) perform temporal modeling based on these extracted visual representations. Finally, a classifier provides the gloss predictions.
\begin{figure}[t]
    \begin{center}
    \includegraphics[width=\linewidth]{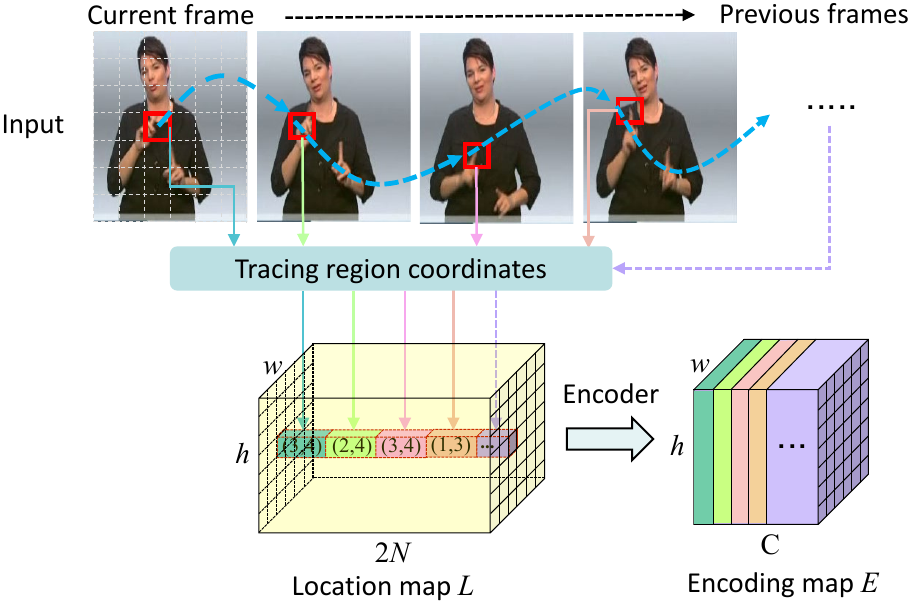}
    \caption{Calculation of location map $L$ and encoding map $E$ in the trajectory module. For each frame, regions are traced back. Coordinates of a region in previous frames are stored at the same location. The location map is passes an encoder to provide encoding map $E$.}
    \label{fig:3}
    \end{center}
\end{figure}
\textbf{TCNet Block}.
We introduce a TCNet block that is responsible for more effective extraction of informative spatio-temporal features. The block operates on subsequences of $N$ frames ($N < T$), and each block represents a stage of feature extractor. The trajectory and correlation modules are processed in parallel (see top of Figure~\ref{fig:2}), after which the feature maps from both modules are aggregated through element-wise multiplication.

\subsection{Trajectory Module}
The trajectory module is motivated by our desire to both capture the trajectories of moving regions, and to apply self-attention to those regions under movement. To this end, we leverage a light-weight motion estimation algorithm \cite{ranjan2017optical} that operates on pairs of input frames to construct an initial location map, which is then encoded. This map provides distinctive information about the movement of the regions. Finally, we obtain the temporal attention map by applying self-attention along the trajectories. The process is illustrated in Figure~\ref{fig:3}.

\textbf{Location Map}.
The location map $L$ will trace back coordinates of regions as the trajectory, which is represented as a group of matrices over time. For an input subsequence of $N$ frames, we obtain $N-1$ backwards optical flow maps $l_t$ ($1 < t \leq N$) by considering pairs of subsequent frames. $l_t$ is a volume of size $w \times h \times 2$, with $w \times h$ the frame resolution and two channels for the x- and y-coordinates. We process the pair of frames in such a way that a coordinate in the optical flow map points to the coordinate in the earlier frame. Specifically, we reverse the order of the pair of frames, and round the floating point optical flow estimates to the nearest coordinates. Each element in $l_t$ thus corresponds to a location in frame $t - 1$. Finally, $l_1$ corresponds to the location indices, typically the pixel coordinates.

We then build the location map $L \in \mathbb{R}^{w \times h \times 2N}$ from all $l_t$. We recursively trace back the location of the region in the previous frame such that each coordinate in $L$ represents the location of the current region in the previous frame. Specially, the coordinate of the same region in previous frames will be recorded at the same location of current frame.

For example, in Figure~\ref{fig:3}, the coordinate of the red region in current frame is (3,4), and the coordinates in the previous three frames are (2,4), (3,4) and (1,3). These coordinate will be written in the same location of the location map (3,4). The vector at each coordinate, through the channel dimension, can be considered a trajectory with alternating x- and y-coordinates over time.

The location map is subsequently encoded through two $1 \times 1$ convolutions to reduce the depth and outputs. The output is encoding map $E$ with the dimensions $w \times h \times C$, with $C$ depending on the stage of the feature extractor. 
In our design, the trajectory generation can be expressed as efficient matrix operations.

\textbf{Temporal Attention Map}.
We first populate a temporally pre-aligned map by changing the coordinates in $L$ by the actual values of the corresponding time frame. In this process, we effectively cancel out gross movement, which allows us to focus on finer-grained movement such as hand signs. For the query in the current frame, we select the keys and values along the trajectories and calculate self-attention to get the feature map related to the trajectories. The calculation of the temporal self-attention volume $TA$ over the input subsequence follows the multi-head scaled dot-product attention \cite{vaswani2017attention}. The generated temporal attention map $TA$ is then added to encoding map $E$, and passed through a $1 \times 1$ convolution as the final result.


\subsection{Correlation Module}
To extract the spatial information, several works~\cite{hu2023continuous,hu2023self} use dilated convolution or self-attention on adjacent frames to emphasize relevant regions. To reduce the computation cost and to reduce the blur introduced from calculating the weighted sum including irrelevant tokens, several works~(e.g., \citet{dong2022cswin,xia2022vision,zhu2023biformer}) propose different sparse attention mechanisms such as local windows~\cite{dong2022cswin}, deformable windows~\cite{xia2022vision}, or use a small sampled subset of key-value pairs for the query~\cite{zhu2023biformer}).

However, for CSLR task, we argue that not all regions in the frame contribute equally, since sign language is mainly conveyed through head and hand regions. Moreover, the correlated regions are dynamic. Hence, it will be more effective if we can design dynamic key-value tokens from correlated regions for each query when calculating self-attention. Based on these observations, we propose a correlation module with a dynamic attention mechanism, which significantly reduces the computation cost and memory use.

\begin{figure}[t]
    \begin{center}
    \includegraphics[width=\linewidth]{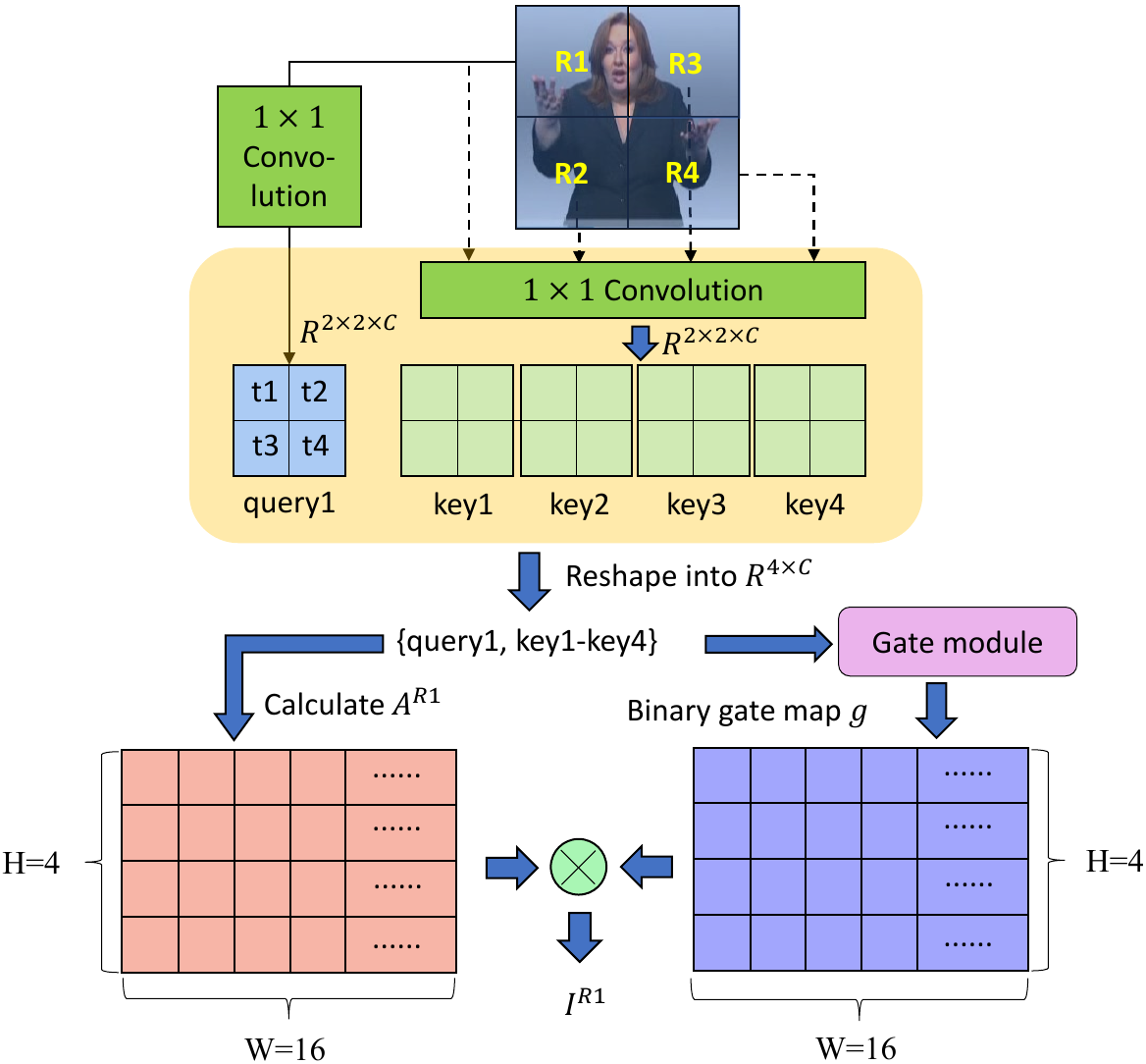}
    \caption{Calculation of the sparse attention matrix $I^{r}$ in the correlation module. In this example, our input is a $4 \times 4$ image and $K = 2$. The frame is split into four regions R1-R4, which pass through a $1 \times 1$ convolution to yield queries and keys. For R1 (other regions analogous), the left branch provides affinity matrix $A^{R1}$, while the right branch generates binary gate map $g$. Both volumes are multiplied element-wisely and pruned to produce sparse attention matrix $I^{R1}$. By combining these results for all regions, we get sparse and dynamic affinity matrix $I^R$ for the current frame.}
    \label{fig:4}
    \end{center}
\end{figure}



Our key idea is to dynamically filter out irrelevant key-value pairs based on the input feature map. To simplify the notations, we discuss the case of single-head self-attention with a single input, although we use multi-head self-attention in practice. The process is shown in Figure~\ref{fig:4}. Given a 2D input feature map $X \in \mathbb{R}^{w \times h \times c}$, we first split it into non-overlapped regions using window size $K$. Each region $X^r \in \mathbb{R}^{K \times K \times c}$ and the corresponding query, key, and value are obtained with linear projections ($1 \times 1 $ convolution):

\begin{equation}
    Q^r = X^rW^q, K^r = X^rW^k, V^r = X^rW^v
\end{equation}
where $W^q,W^k,W^v \in \mathbb{R}^{1 \times 1 \times c}$ are projection weights for the query, key and value, respectively.

For region-level queries $Q^r$ and keys $K^r$, the region-to-region affinity matrix, $A^r$, can be obtained via matrix multiplication between $Q^r$ and the transposed $K^r$: 
\begin{equation}
    A^r = Q^r(K^r)^T
\end{equation}

An example of calculating the regional affinity matrix $A^{R1}$ for Region 1 is shown in Figure~\ref{fig:4}. The value of every entry in affinity matrix $A^r$ measures how much two tokens are semantically related. For example, in Figure~\ref{fig:4}, the first line indicates the affinity of token $t1$ to tokens from key1, key2, key3, and key4. To reduce the computation cost and to reduce blur introduced by having irrelevant tokens contribute to the weighted sum, the core step that we perform next is to prune the affinity graph by dynamically keeping $k$ connections for each region. To realize this, we use a gate module to generate a binary gate map $g$. Through element-wise multiplication with $A^r$, we obtain a sparse matrix $I^r$. Hence, the $i^{th}$ row of $I^r$ contains $k$ indices of the most relevant regions for the $i^{th}$ region. Since the generated $g$ depends on the input frames, the output $g$ is fully dynamic.

For a region with $\{query1, key1, key2, key3, key4\} \in \mathbb{R}^{K^2 \times c}$, the input of the gate module is calculated as the concatenation of the vector product of $query1$ with each of the four transposed keys.

The gate module consists of two $3 \times 3$ convolutions with a padding and stride of one. The module outputs an initial gate map $g'$ that has the same size as $A^r$. We generate binary gate map $g$ from $g'$ by pruning the affinity graph. To make sure we can back-propagate the error through the discrete $g$, we adopt Improved SemHash~\cite{kaiser2018discrete}. We define the function to get the binary gate feature map $g$ as follows:

\begin{equation}
    g_\alpha = \sigma'(g') \quad and \quad g_\beta = 1(g'>0)
\end{equation}
where $\sigma'(x)$ is the saturating sigmoid function~\cite{kaiser2016can,kaiser2015neural}:
\begin{equation}
    \sigma'(x) = max(0, min(1, 1.2\sigma(x)-0.1))
\end{equation}
with sigmoid function $\sigma$.

\begin{table*}[t]
\begin{center}
\small
\begin{tabular}{lccccccccc}
\toprule
\multirow{3}{*}{\textbf{Method}}  & \multirow{3}{*}{\textbf{Backbone}}  & \multirow{3}{*}{\textbf{FLOPs}} & \multirow{3}{*}{\textbf{Params}} & \multicolumn{4}{c}{\textbf{PHOENIX14}} &  \multicolumn{2}{c}{\textbf{PHOENIX14-T}} \\
 &   & & &\multicolumn{2}{c}{DEV}&\multicolumn{2}{c}{TEST}   & DEV&  TEST\\
 &    &  & &\textbf{del / ins} & WER & \textbf{del / ins} &  WER & WER & WER \\
\midrule
SFL~\cite{niu2020stochastic}&  ResNet18 &1180.2G &33.7M &7.9 / 6.5  & 26.2 &7.5 / 6.3 &26.8 &25.1 &26.1 \\
FCN~\cite{cheng2020fully} &  Custom &983.0G& 20.3M  & 7.8 / 4.7 & 23.7 &7.2 / 4.5 &23.9 &23.3 &25.1 \\
CMA~\cite{pu2020boosting}   &  GoogLeNet &- &- & 7.3 / 2.7  & 21.3 &7.3 / 2.4 &21.9 &- &- \\
VAC~\cite{min2021visual} &  ResNet18 &1120.4G &22.3M & 7.9 / 2.5  & 21.2 &8.4 / 2.6 &22.3 &21.4 &23.9 \\
SMKD~\cite{hao2021self}&  ResNet18 &1032.3G &20.3M & 6.8 / 2.5  & 20.8 &6.3 / 2.3 &21.0 &20.8 &22.4 \\
TLP~\cite{hu2022temporal}    & ResNet18 &2950.5G
&48.0M & 6.3 / 2.8  & 19.7 &6.1 / 2.9 &20.8 &19.4 &21.2 \\
SEN~\cite{hu2023self}  & ResNet18 &1144.2G &23.1M & 5.8 / 2.6  & 19.5 &7.3 / 4.0 &21.0 &19.3 &20.7 \\
CorrNet~\cite{hu2023continuous}  & ResNet18 &1035.4G &20.5M & 5.6 / 2.8  & 18.8 &5.7 / 2.3 &19.4 &18.9 &20.5\\
\midrule
SLT*~\cite{camgoz2018neural}  & GoogLeNet &- &- & -  & - &- &- &24.5 &24.6\\
HMM*~\cite{koller2019weakly}  & GoogLeNet &1488.2G &44.0M & 8.1 / 3.8 & 26.0 &8.2 / 3.6 &26.0 &22.1 &24.1\\
DNF*~\cite{cui2019deep}  & GoogLeNet &1487.3G &43.30M & 7.3 / 3.3  & 23.1 &6.7 / 3.3 &22.9 &22.7 &24.0\\
STMC*~\cite{zhou2020spatial}  & VGG11 &1742.1G &40.71M & 7.7 / 3.4  & 21.1 &7.4 / 2.6 &20.7 &19.6 &21.0\\
C$^2$\text{SLR}*~\cite{zuo2022c2slr}  & ResNet18 &1124.6G &32.6M & 6.8 / 3.0  & 20.5 &7.1 / 2.5 &20.4 &20.2 &20.4\\
\midrule
\textbf{TCNet (ours)}& ResNet18 &932.1G &19.3M & \textbf{5.5 / 2.4}  & \textbf{18.1} & \textbf{5.4 / 2.0} & \textbf{18.9} & \textbf{18.3} & \textbf{19.4}\\
\bottomrule
\end{tabular}
\end{center}
\caption{Comparison with state-of-the-art on PHOENIX14 and PHOENIX14-T. WER in \%. Networks with $\ast$ are trained using additional inputs such as face or hand features, or pre-extracted heatmaps. Best results in bold.}
\label{tab:2}
\end{table*}

Here, $g_\alpha$ is a real-valued gate map with all entries falling in the interval $[0.0, 1.0]$, while $g_\beta$ is a binary gate map. We observe that $g_\beta$ has the desirable binary property that we want to use in our model, but the gradient of $g_\beta$ w.r.t $g'$ is not defined. On the other hand, the gradient of $g_\alpha$ w.r.t $g'$ is well defined, but $g_\alpha$ is not binary. In forward propagation, we randomly use $g = g_\alpha$ for half of the training samples and use $g = g_\beta$ for the remaining samples. When $g_\beta$ is used, we follow the solution in~\cite{kaiser2018discrete} and define the gradient of $g_\beta$ w.r.t $g'$ to be the same as the gradient of $g_\alpha$ w.r.t $g'$ in the backward propagation. In the evaluation and inference phases, we use $g = g_\beta$.

\subsection{Classifier}
The final classifier is essentially a fully connected layer. It performs classification based on temporal features extracted by the Bi-LSTM layer. The output of the Bi-LSTM layer is a 2D tensor with dimensions (batch size, hidden size), where hidden size is the size of the LSTM hidden states. In practice, we initialize the classifier with Xavier Uniform initialization. The output of the Bi-LSTM layer is multiplied with the weight matrix to generate the final output.

\section{Experiments}
\label{sec:experiments}

We provide extensive experiments to evaluate the effectiveness of our method. Datasets and evaluation metrics are introduced first. Then, we detail the experimental setup and analyze the results.

\subsection{Dataset and Metrics}
We conduct our experiments on four public datasets.

\textbf{PHOENIX14}~\cite{KOLLER2015108} is recorded from a German weather forecast broadcast with nine actors before a clean background with a resolution of $210 \times 260$. It contains 6,841 sentences with a vocabulary of 1,295 signs, divided into 5,672 training, 540 development (DEV), and 629 testing (TEST) samples.

\textbf{PHOENIX14-T}~\cite{camgoz2018neural} covers the same domain as PHOENIX14 and is available for both CSLR and sign language translation tasks. It has a vocabulary of 1,085 signs and contains 8,247 sentences. These are divided over training (7,096 instances), development (DEV, 519 samples), and testing (TEST, 642 samples) sets.

\textbf{CSL}~\cite{huang2018video} is collected in the laboratory environment by fifty signers with a vocabulary size of 178 with 100 sentences. It contains 25,000 videos, divided into training and testing sets by a ratio of 4:1.

\textbf{CSL-Daily}~\cite{zhou2021improving} revolves the daily life, recorded indoors at 30fps by 10 signers. CSL-Daily has the largest vocabulary size (2k) among the tested datasets, covering a broad range of topics. It contains 20,654 sentences, divided into 18,401 training samples, 1,077 development (DEV) samples and 1,176 samples for testing (TEST).

In continuous SLR, word error rate (WER) is the most widely-used metric to evaluate the performance. WER is essentially an edit distance and reflects the lowest number of \textbf{sub}stitution, \textbf{ins}ertion, and \textbf{del}etion operations to transform the predicted sentence into the reference sequence:
\begin{equation}
    \text{WER}=\frac{\text{\#sub + \#ins + \#del}}{\text{length of reference}}
\end{equation}

\subsection{Training Details}
Following previous works, ours uses a ResNet-18 backbone. We train the network using the following loss function:

\begin{equation}
    \mathcal{L} = \mathcal{L}_{CTC}+\mathcal{L}_{VE}+\mathcal{L}_{VA}+\frac{||G(X)||_1}{c}
    \label{eqn:loss}
\end{equation}
where $\mathcal{L}_{CTC}$ is the popular CTC loss ~\cite{graves2006connectionist} to predict the probability of the target gloss, and $\mathcal{L}_{VE}$ and $\mathcal{L}_{VA}$ are VE and VA losses from VAC~\cite{min2021visual}. The final term is a L1 regularization term to encourage $g$ to be sparse. Here, $c$ is the size of $g$. The backbone network receives no gradients from the second term, while the gate module receives gradients from the first two terms in Equation~\ref{eqn:loss}.

In the trajectory module, we use the pre-trained SPyNet~\cite{ranjan2017optical} as motion estimation network and we set size of the subsequences to $N = 12$. An ablation study is performed later. The number of hidden states in the 2-layer Bi-LSTM encoder is set to be 1,024, followed by a fully connected layer for gloss prediction. We train our model for 80 epochs using the Adam optimizer. The initial learning rate is set to be $1 \times 10^{-4}$ decayed by 5 after 40 and 60 epochs, and the weight decay is set to $5 \times 10^{-3}$. All input frames are first resized to $256\times 256$, and then randomly cropped to $224\times 224$ with 50\% horizontal flipping and 20\% temporal rescaling during training. During inference, a $224\times 224$ center crop is used.

\subsection{Comparison with State-of-the-Art}
We first compare TCNet to the current state-of-the-art.

\textbf{PHOENIX14 and PHOENIX14-T}. A comparison between TCNet and current state-of-the-art methods on PHOENIX14 and PHOENIX14-T appears in Table~\ref{tab:2}. We split the table into three parts. The middle part, works annotated with $\ast$, contains results that are obtained using additional inputs like face or hand features.
 
TCNet outperforms state-of-the-art methods by a clear margin on both datasets. Compared to methods without additional inputs, TCNet obtains higher performance than the second best CorrNet on both PHOENIX14 (18.8$\rightarrow$18.1 and 19.4$\rightarrow$18.9 on DEV and TEST, respectively) and PHOENIX14 (18.9$\rightarrow$18.3 and 20.5$\rightarrow$19.4 on DEV and TEST). Compared to methods with additional inputs, our TCNet still achieves better performance, since the advanced design of the correlation module can dynamically extract correlated information from face or hand regions, and the trajectory module can trace the trajectories of these regions. For example, compared to C$^2$\text{SLR}, with the same backbone, TCNet lowers the WER by 2.4\% and 1.5\% on PHOENIX14 DEV and TEST, respectively (20.5$\rightarrow$18.1, 20.4$\rightarrow$18.9). Although C$^2$\text{SLR} outperforms CorrNet by 0.1\% on PHOENIX14-T TEST, TCNet surpasses C$^2$\text{SLR} by a 1.0\%(20.4$\rightarrow$19.4) WER reduction. We see the same pattern for the number of deletions and insertions.

TCNet has the lowest number of FLOPs and trained parameters of all the tested networks. These include the additional operations and parameters of the motion estimation module. The improved performance therefore is not due to the increased model capacity.
 
\textbf{CSL}. Table~\ref{tab:3} shows that our TCNet achieves the new state-of-the-art performance, lowering the previous WER of 0.8 ~\cite{hu2023continuous,hu2023self} to 0.7. Again, TCNet has the lowest number of FLOPs and parameters from the tested methods.

\begin{table}[htb]
\begin{center}
\small
\begin{tabular}{lccc}
\hline

\textbf{Method} & \textbf{FLOPs} &\textbf{Param.} & \textbf{WER} \\
\toprule
SF-Net~\cite{yang2019sf}&1056.1G & 27.9M & 3.8\\
FCN~\cite{cheng2020fully}&977.2G & 20.3M & 3.0\\
STMC~\cite{zhou2020spatial}&1726.1G &40.71M&2.1\\
VAC~\cite{min2021visual}& 1107.4G & 22.3M & 1.6\\
C$^2$\text{SLR}~\cite{zuo2022c2slr} &1107.2G &31.7M & 0.9\\
CorrNet~\cite{hu2023continuous} & 1033.1G & 20.3M & 0.8\\
SEN~\cite{hu2023self} &1136.1G &23.0M & 0.8\\
\textbf{TCNet (ours)}& 929.8G& 18.8M &\textbf{0.7} \\
\bottomrule
\end{tabular}
\end{center}
\caption{Comparison with state-of-the-art methods on CSL. WER in \%. Best results in bold.}
\label{tab:3}

\end{table}

\textbf{CSL-Daily}.
Table~\ref{tab:4} shows that TCNet also outperforms all state-of-the-art works upon on this challenging dataset. In particular, we reduce the previous best TEST WER of CorrNet~\cite{hu2023continuous} by 0.8\%. These results demonstrate the merits of TCNet to address a more general input domain.

\begin{table}[htb]
\begin{center}
\small
\begin{tabular}{lcc}
\hline
\textbf{Method} & \textbf{DEV} & \textbf{TEST} \\
\hline
LS-HAN~\cite{huang2018video} &39.0 &39.4\\
TIN~\cite{cui2019deep} &32.8 &32.4\\
Joint-SLRT~\cite{camgoz2020sign} &33.1 &32.0\\
FCN~\cite{cheng2020fully} &33.2 &32.5\\
BN-TIN~\cite{zhou2021improving} &33.6 &33.1\\
SEN~\cite{hu2023self} &31.1 &30.7\\
CorrNet~\cite{hu2023continuous}  & 30.6 & 30.1\\
\textbf{TCNet (ours)}&\textbf{29.7} & \textbf{29.3}\\
\hline
\end{tabular}
\end{center}
\caption{Comparison with state-of-the-art methods on CSL-Daily. WER in \%. Best results in bold.}
\label{tab:4}
\end{table}

\subsection{Ablation Study}
To analyze the merits of the TCNet block, we perform ablations. First, we assess the contribution of the trajectory and correlation modules individually on the performance. Second, we investigate the influence of the backbone. Third, we experiment with various combination operations to combine results from the two modules. Fourth, we show the performance with input subsequences of various lengths $N$.

\textbf{Influence of Trajectory and Correlation Modules}. We assess the contribution of the two modules separately using the same experimental setting on PHOENIX14. The results are summarized in Table~\ref{tab:6}. Compared with the baseline, the trajectory module and correlation module each bring a notable reduction in WER in the range 1.0-1.6\%. When both modules are used together, there is a further reduction of 0.5-1.1\%. This shows that the two modules are both effective, and their influence is partly complementary.

\begin{table}[b]
\begin{center}
\small
\begin{tabular}{cccc}
\hline

\textbf{Trajectory} & \textbf{Correlation} & \textbf{DEV WER}  & \textbf{TEST WER} \\
\hline
 & &20.2 &21.0\\
 &\Checkmark &18.8 (-1.4) &19.8 (-1.2) \\
\Checkmark &\ &19.2 (-1.0) &19.4 (-1.6) \\
\Checkmark &\Checkmark&\textbf{18.1} (-2.1) & \textbf{18.9} (-2.1) \\
\hline
\end{tabular}
\end{center}
\caption{Influence of the trajectory and
correlation modules on PHOENIX14.  WER in \%. Best results in bold.}
\label{tab:6}
\end{table}

\textbf{Influence of Backbone.} Instead of ResNet-18, we incorporate TCNet blocks in backbones GoogLeNet~\cite{szegedy2015going}, VGG11~\cite{simonyan2014very}, SqueezeNet~\cite{hu2018squeeze}, and ShuffleNet V2~\cite{ma2018shufflenet}. The aggregated result of trajectory and correlation modules are added with three spatial downsampling layers in each backbone. Using the same experimental procedure as before, we summarize results on PHOENIX14 in Table~\ref{tab:5}. We observe that the use of TCNet blocks consistently improves the performance of backbones with a reduction in WER in the range 1.8-2.6\%. These results demonstrate the choice of backbone matters, but the inclusion of the trajectory and correlation modules is always beneficial.

\begin{table}[t]
\begin{center}
\small
\begin{tabular}{lcc}
\hline
\textbf{Backbone} & \textbf{DEV WER} & \textbf{TEST WER} \\
\hline
SqueezeNet &22.2 &22.6\\
\textbf{+TCNet}&\textbf{19.8} (-2.4) & \textbf{20.0} (-2.6) \\
\hline
ShuffleNet V2 &21.7 &22.2\\
\textbf{+TCNet}&\textbf{19.4} (-2.3) & \textbf{20.0} (-2.2) \\
\hline
GoogLeNet &21.4 &21.5\\
\textbf{+TCNet}&\textbf{19.0} (-2.4) & \textbf{19.3} (-2.2) \\
\hline
VGG11 &20.7 &21.0\\
\textbf{+TCNet}&\textbf{18.7} (-2.0) & \textbf{19.2} (-1.8) \\
\hline
\end{tabular}
\end{center}
\caption{Backbones with and without TCNet blocks on PHOENIX14. WER in \%. Best results in bold.}
\label{tab:5}
\end{table}

\textbf{Combination of Trajectory and Correlation Modules.}
We experiment with alternative operators to aggregate the results from the trajectory and correlation modules: summation and concatenation. Comparisons with the element-wise multiplication are presented in Table~\ref{tab:7}. The difference in performance is modest, with only 0.1-0.4\% higher WER for alternative ways of combining the module outputs.

\begin{table}[t]
\begin{center}
\small
\begin{tabular}{lcc}
\hline
\textbf{Operation} & \textbf{DEV WER (\%)} & \textbf{TEST WER (\%)} \\
\hline
Concatenation& 18.5 & 19.2\\
Summation & 18.3 & 19.0\\
Multiplicatioin & \textbf{18.1}&\textbf{18.9}\\
\hline
\end{tabular}
\end{center}
\caption{Alternative combination operators of trajectory and correlation modules on PHOENIX14. Best results in bold.}
\label{tab:7}
\end{table}

\textbf{Number of Frames for Tracing the Trajectory.}
In the trajectory module, we trace the trajectories in subsequences of length $N$. Since the number of number may affect the effectiveness of the temporal information encoding, we investigate the performance on PHOENIX14 for various $N$. We summarize the results in Table~\ref{tab:8}. The best performance is achieved for $N=12$, approximately half a second, our default setting.

\begin{table}[t]
\begin{center}
\small
\begin{tabular}{cccccccc}
\hline
\textbf{Frames} & \textbf{1} & \textbf{2} & \textbf{4} & \textbf{8} & \textbf{12} & \textbf{16} & \textbf{18} \\
\hline
DEV & 18.8 &18.7&18.3&18.2&18.1&18.9&19.3\\
TEST &19.5 &19.5&19.2&19.0&18.9&19.5&20.0\\
\hline
\end{tabular}
\end{center}
\caption{Subsequence length $N$ in the trajectory module on PHOENIX14. WER in \%. Best results in bold.}
\label{tab:8}
\end{table}

\begin{figure}[htb]
    \begin{center}
    \includegraphics[width=\linewidth]{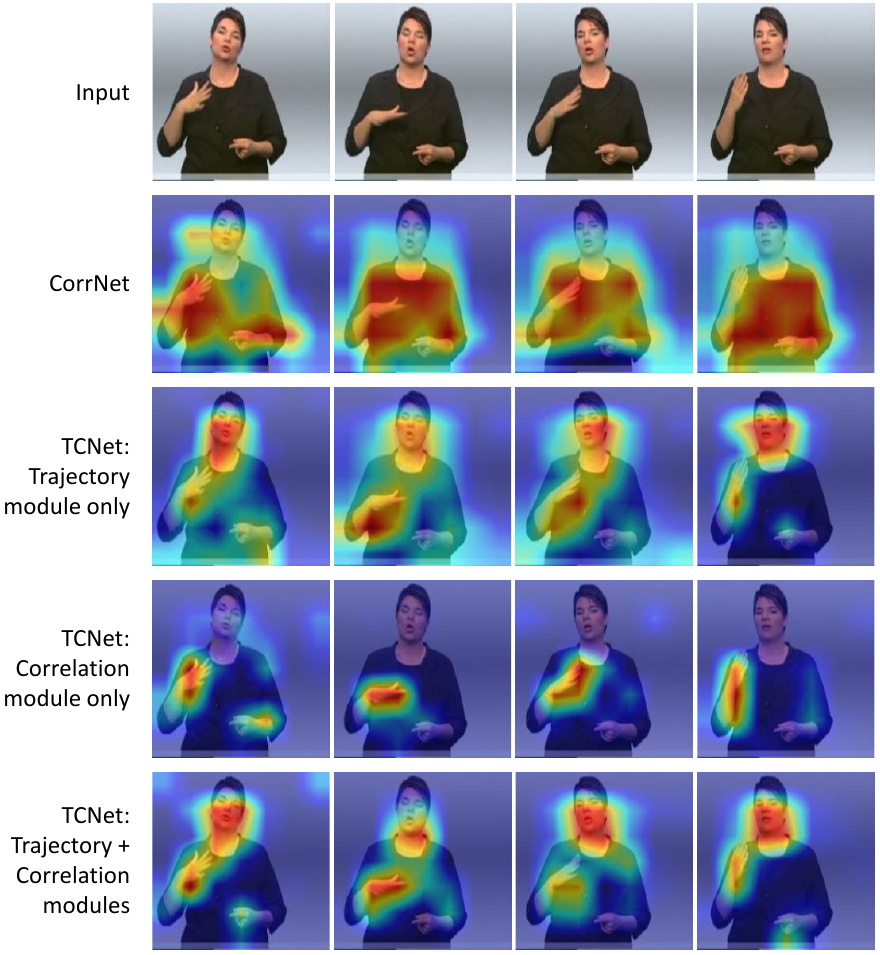}
    \caption{Grad-CAM heatmaps for CorrNet and TCNet with only the Trajectory module, only the Correlation module, and with both modules.}
    \label{fig:5}
    \end{center}
\end{figure}

\subsection{Visualizations}
In Figure~\ref{fig:5}, we show Grad-CAM~\cite{selvaraju2017grad} heatmaps for CorrNet~\cite{hu2023continuous} and TCNet with either and both trajectory and correlation module in the TCNet block. While CorrNet is relatively unfocused and also attends to irrelevant regions, TCNet focuses almost exclusively on the face and right hand that performs the gestures. 

The trajectory module mainly focuses on the head and right hand regions of the signer since the body movement mainly happens in these regions, while the left hand remains almost stationary. Overall, the attention is not too narrow. In contrast, the correlation module focuses mostly on the right hand, a bit on the left hand but not on the face. The two modules combined demonstrate attention for the most informative regions to understand the sign.

\section{Conclusion}
\label{sec:conclusions}
In this work, we have addressed the extraction of more informative spatio-temporal features for continuous sign language recognition (CSLR). We have proposed novel trajectory and correlation modules that are incorporated in a hybrid network: TCNet. The trajectory module traces movements of regions, mainly of the hands and face, over time to allow for self-attention along the trajectory. The correlation module dynamically filters out irrelevant key-value pairs. Based on the aggregation of both modules, we have demonstrated state-of-the-art results on common CSLR datasets. For example, we improve over the previous state-of-the-art \cite{hu2023continuous} by 1.5\% and 1.0\% word error rate on PHOENIX14 and PHOENIX14-T, respectively. Ablation studies validate the contribution of both modules, and the robustness of the results with different design choices.

\bibliography{aaai24}

\end{document}